\documentclass[]{spie}  

\usepackage{amsmath,amsfonts,amssymb}
\usepackage{graphicx}
\usepackage[colorlinks=true, allcolors=blue]{hyperref}

\newcommand\blfootnote[1]{%
	\begingroup
	\renewcommand\thefootnote{}\footnote{#1}%
	\addtocounter{footnote}{-1}%
	\endgroup
}

\title{CT synthesis from MR images for orthopedic applications in the lower arm using a conditional generative adversarial network}

\author[a]{Frank Zijlstra}
\author[b]{Koen Willemsen}
\author[a]{Mateusz C. Florkow}
\author[b]{Ralph J.B. Sakkers}
\author[b]{Harrie H. Weinans}
\author[b]{Bart C.H. van der Wal}
\author[a,c]{Marijn van Stralen}
\author[a,c]{Peter R. Seevinck}

\affil[a]{Image Sciences Institute, University Medical Center Utrecht, Utrecht, The Netherlands}
\affil[b]{Department of Orthopedics, University Medical Center Utrecht, Utrecht, The Netherlands}
\affil[c]{MRIguidance B.V., Utrecht, The Netherlands}

\begin{document}
\maketitle

\blfootnote{This work has been accepted at the SPIE Medical Imaging 2019, Image Processing conference, paper 10949-54.}

\begin{abstract}
\textbf{Purpose:} To assess the feasibility of deep learning-based high resolution synthetic CT generation from MRI scans of the lower arm for orthopedic applications.

\noindent\textbf{Methods:} A conditional Generative Adversarial Network was trained to synthesize CT images from multi-echo MR images. A training set of MRI and CT scans of 9 ex vivo lower arms was acquired and the CT images were registered to the MRI images. Three-fold cross-validation was applied to generate independent results for the entire dataset. The synthetic CT images were quantitatively evaluated with the mean absolute error metric, and Dice similarity and surface to surface distance on cortical bone segmentations.

\noindent\textbf{Results:} The mean absolute error was 63.5 HU on the overall tissue volume and 144.2 HU on the cortical bone. The mean Dice similarity of the cortical bone segmentations was 0.86. The average surface to surface distance between bone on real and synthetic CT was 0.48 mm. Qualitatively, the synthetic CT images corresponded well with the real CT scans and partially maintained high resolution structures in the trabecular bone. The bone segmentations on synthetic CT images showed some false positives on tendons, but the general shape of the bone was accurately reconstructed.

\noindent\textbf{Conclusions:} This study demonstrates that high quality synthetic CT can be generated from MRI scans of the lower arm. The good correspondence of the bone segmentations demonstrates that synthetic CT could be competitive with real CT in applications that depend on such segmentations, such as planning of orthopedic surgery and 3D printing.
\end{abstract}


\section{Introduction}
Various studies have demonstrated the feasibility of generating clinically relevant synthetic Computed Tomography (CT) images from conventional Magnetic Resonance (MR) images in radiotherapy for dose calculations \cite{dinkla_dosimetric_2018} and in PET/MRI for attenuation correction \cite{liu_deep_2017}. Since the synthetic CT is intrinsically registered to the MR images, volumes of interest can be segmented on the MRI scan and directly applied to the CT/PET images without registration errors. Furthermore, an MR-only workflow has a lower cost than a CT-MR fusion workflow and does not deposit ionizing radiation.

The primary difficulty in synthetic CT generation is the accurate depiction of cortical bone. Cortical bone appears as a signal void on conventional MRI sequences due to its short $T_2$ value. Other structures may also appear as a signal void on MR images, such as air, tendons, ligaments, blood vessels, and boundaries between water and fat. Consequently, distinguishing bone from other signal voids requires contextual information. The current state of the art methods for synthetic CT generation are based on deep convolutional neural networks (CNNs) \cite{han_mr-based_2017,nie_medical_2017,wolterink_deep_2017}, which can automatically learn to detect complex features, such as texture and shape context, to help distinguish bone from other signal voids. Using these deep CNNs, MR images can be directly translated into CT Hounsfield units (HU).

Because of the difficulties in depicting cortical bone with MRI, the use of MRI has been limited in orthopedic applications that require accurate bone segmentations or assessment of bone density. Instead, CT is a natural choice of imaging in orthopedic applications because of its sensitivity to bone density. Examples of applications of CT in orthopedics include diagnosis and surgical planning in complex fractures \cite{bahrs_indications_2009}, assessment of bone mineral density \cite{schreiber_use_2014}, planning of osteotomies \cite{dobbe_computer-assisted_2011}, range of motion analysis \cite{colaris_three-dimensional_2014}, and 3D printing \cite{auricchio_3d_2016}. In most of these applications, CT scans are used to obtain high quality segmentations of bone. Synthetic CT could potentially replace real CT in these type of orthopedic applications. Especially in the pediatric population, the lack of ionizing radiation in MRI scans would be a benefit. Furthermore, MRI offers information about soft tissues that may be helpful in orthopedic applications, such as the location and possible pathology of cartilage, tendons, and muscles.

In this study, we assess the feasibility and accuracy of high resolution synthetic CT generation from MRI scans of the lower arm for orthopedic purposes. Since bone segmentation is an important part of clinical workflows, we focused our evaluation of the synthetic CT images on bone segmentation and visualization. For synthetic CT generation we trained a 2D conditional generative adversarial network (cGAN), which has been shown to be effective in paired image-to-image translation \cite{isola_image-to-image_2016}. This study was limited to an ex vivo dataset to limit the impact of registration errors on the results. As such, this study assesses the quality of synthetic CT that can be expected to be generated from MR images on the condition that a high quality paired training set is available.

\section{Methods}

\subsection{Network}
We used a conditional Generative Adversarial Network (cGAN) \cite{isola_image-to-image_2016} with L1 loss to perform translation from 2D multi-echo MR images to CT images. The generator network was based on a UNet architecture \cite{ronneberger_u-net_2015} with 4 image scales. The top level layers had 32 channels, which was doubled for every lower scale. Downsampling was implemented with average pooling and upsampling was implemented with nearest neighbor interpolation, which forces the network to only consider lower spatial resolution information in the lower scales. On each scale there were four convolution layers, each of which was followed by a pixelwise normalization and ReLU activation. Pixelwise normalization explicitly normalizes each pixels feature vector to be zero mean and unit standard deviation during both training and testing. The discriminator network was implemented as a 4-layer CNN with 64 channels in each layer. The convolution kernel size in both the generator and discriminator networks was $3 \times 3$, resulting in a receptive field of $124 \times 124$ for the generator and $9 \times 9$ for the discriminator.

\begin{figure}
	\begin{center}
		\includegraphics[width=\textwidth]{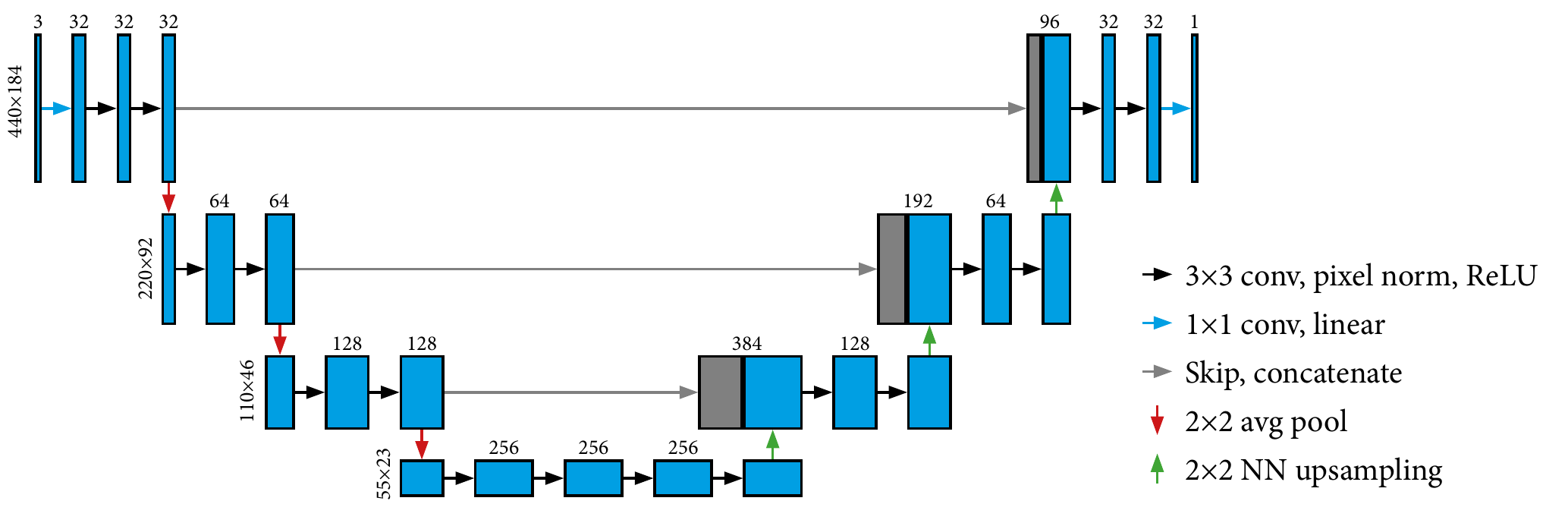}
	\end{center}
	
	\caption{UNet network architecture. The network consists of 4 scales, with 4 convolution layers on each scale. The first scale has two additional convolution layers to transform the 3 input channels to 32 ($1\times1$ conv + ReLU activation), and to transform the final 32 channels to 1 output channel ($1\times1$ conv + linear activation). Downsampling is performed with average pooling, and upsampling is performed with nearest neighbour (NN) interpolation.}
	\label{fig_network}
\end{figure}

\subsection{Dataset}
For training of the cGAN we acquired a paired MR and CT dataset of 9 ex vivo adult human arms, which were fixated in an extended and pronated position to allow both MR and CT scanning with relatively little movement and deformation in between scans. The MR scan was a multi-echo gradient echo scan acquired at a field strength of 3 tesla (Ingenia, Philips Healthcare, Best, The Netherlands) with the following scan parameters: 1.2 mm isotropic resolution (reconstructed to 0.6 mm), $313 \times 103 \times 128$ mm FOV, echo times 2.1/3.25/4.4 ms, repetition time 6.9 ms, flip angle 15, and a total scan duration of 151 seconds. CT scans were obtained at a resolution of $0.3 \times 0.3$ mm and a slice thickness of 0.4 mm.

\begin{figure}
\begin{center}
	\includegraphics[width=\textwidth]{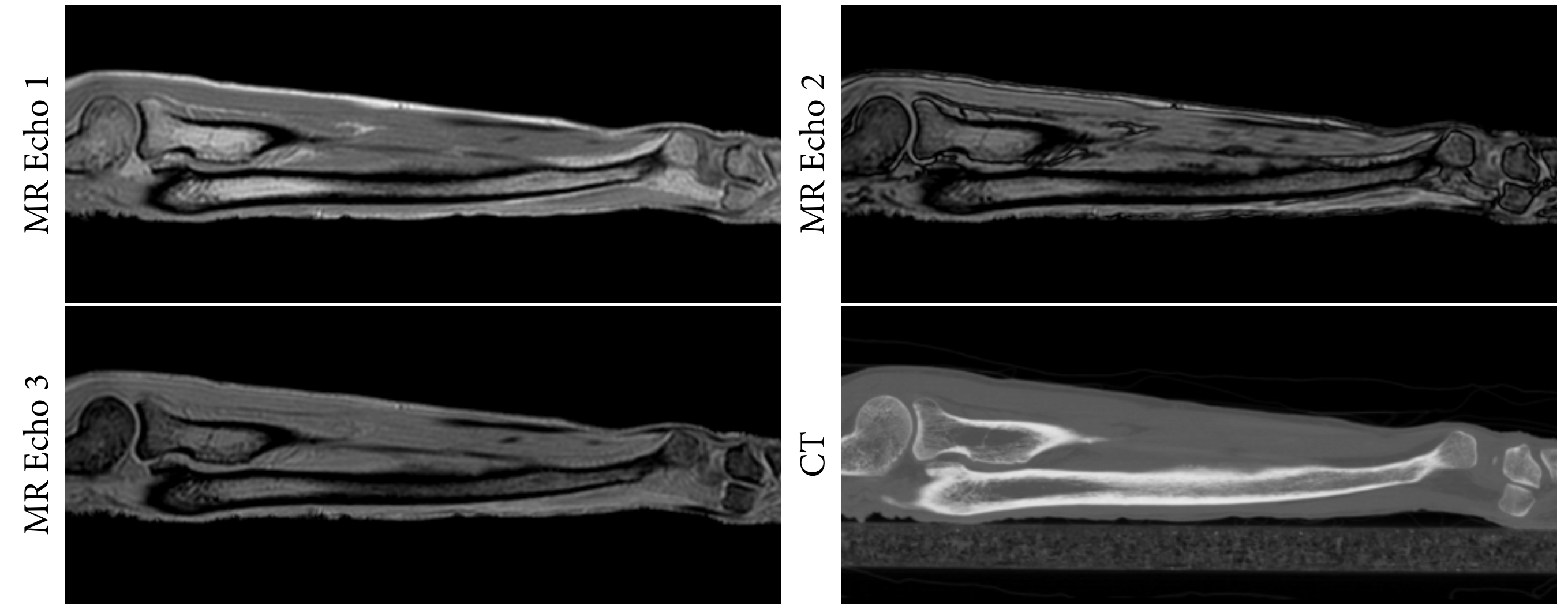}
\end{center}

\caption{Sample of the acquired dataset for subject 1: three echoes from the MR scan and the registered CT image.}
\label{fig_images}
\end{figure}

The CT scans were rigidly registered to the MR scans using the iterative closest point algorithm operating on the point sets of high values on the CT scans and low values on the MR scans, which indicate the presence of cortical bone. After registration, all scans were cropped to the overlapping area between the MR and registered CT images, while maintaining a fixed image size for the entire dataset ($440 \times 184 \times 213$ voxels). Figure \ref{fig_images} shows the acquired MR and CT images for one subject.

\subsection{Training}
The CNN was trained with the Adam optimizer \cite{kingma_adam_2015} for 500 epochs with a batch size of 4. The learning rate was fixed to 0.0002 for the first 400 epochs, and linearly reduced to 0 during the final 100 epochs. Random x- and y-shifts up to 8 voxels were applied during training for data augmentation. The dataset was split into 3 folds for cross-validation, which allowed evaluation of synthetic CT on the entire dataset. This resulted in a training set of 6 scans, of which 4 scans were used for training and 2 scans were used for validation, and a test set of 3 scans for each cross-validation fold. The training time for each fold was approximately 14 hours on an Nvidia Quadro P6000 GPU. Generation of the synthetic CT images took approximately 5 seconds per complete volume.

\subsection{Evaluation}
The resulting synthetic CT images were quantitatively evaluated using a Mean Absolute Error (MAE) metric on both the entire tissue volume and the bone segmentation, the Dice similarity index \cite{dice_measures_1945} of cortical bone segmentations, and the surface to surface distance between bone surfaces extracted from synthetic and real CT images. The Dice similarity and surface-distance metrics are more sensitive to voxel-level accuracy of the bones than global metrics, such as MAE, used in previous studies. Cortical bone segmentations were obtained by applying a threshold of 200 HU to the images. A 200 HU isosurface was calculated for both the real and synthetic CT to perform surface to surface distance measurements. The surface distance was calculated per vertex on the real CT isosurface as the closest distance to a vertex on the synthetic CT isosurface, and vice versa.

\section{Results}
\begin{figure}
	\begin{center}
		\includegraphics[width=\textwidth]{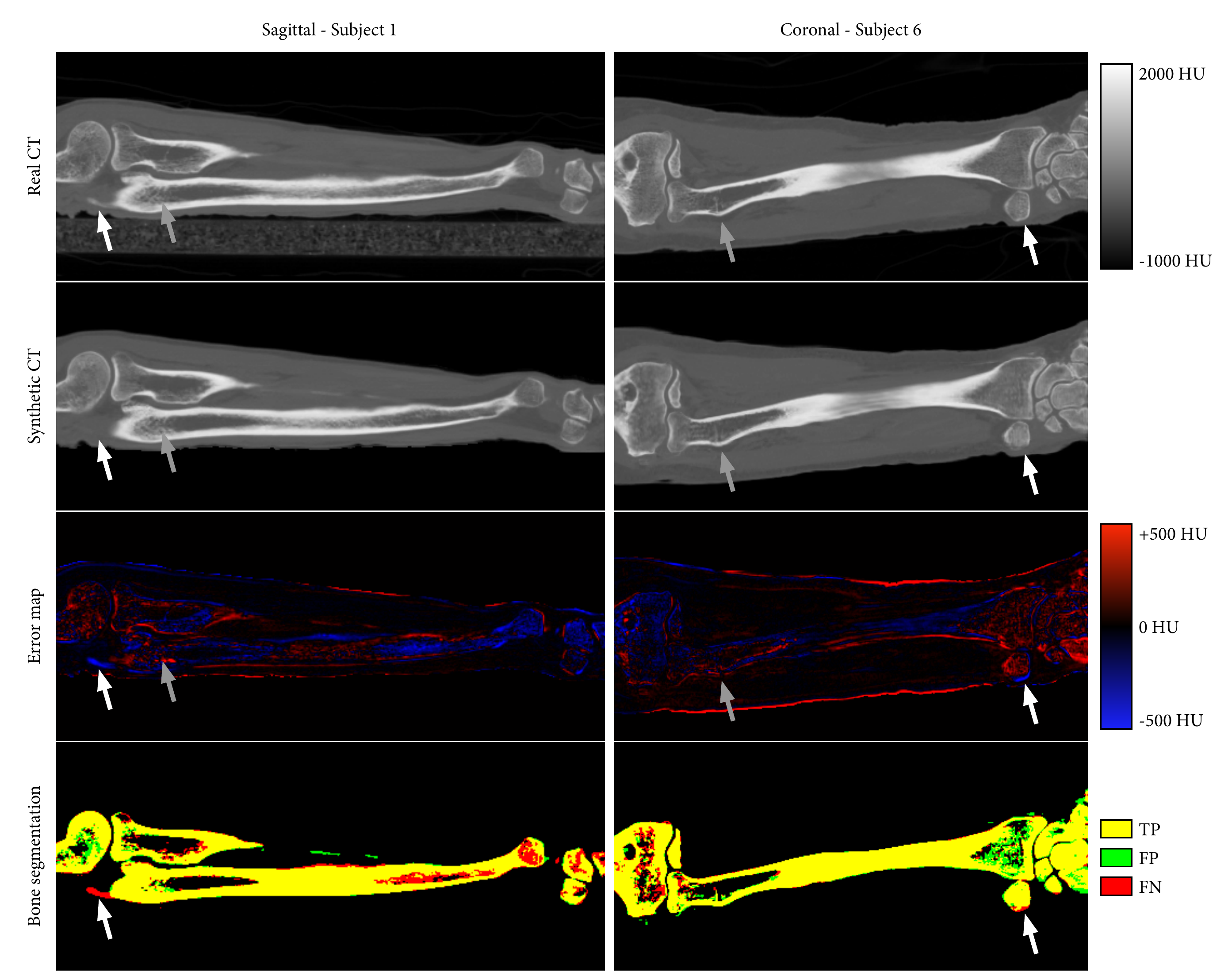}
	\end{center}
	
	\caption{Synthetic CT compared to real CT on a sagittal and a coronal slice for two subjects. Error maps highlight the differences between the two images, with blue values indicating areas where the synthetic CT had lower intensity than the real CT and red values indicating higher intensity. Bone segmentations ($>200$ HU) for both synthetic CT and real CT were overlaid and marked as true positive (TP, yellow), false positive (FP, green), and false negative (FN, red). White arrows indicate examples of regions with large errors in the bone shape. Gray arrows indicate examples of regions where internal bone structure is preserved by the synthetic CT.}
	\label{fig_pct}
\end{figure}

Figure \ref{fig_pct} shows synthetic CT images as generated by our approach for two subjects. The synthetic CT images qualitatively have a good correspondence to the real CT images. In general the shape of the bones is well-represented in the synthetic CTs. Two major errors were marked with white arrows, one in the proximal end of the ulna on the sagittal slice and one in the distal end of the ulna on the coronal slice. The error maps spuriously show some additional errors on the boundary of the arm and on edges of bone, which are due to residual registration errors. While the internal trabecular structure of the bones has slightly more detail on the real CT, these structures are still present on the synthetic CT, like indicated on the radius in both the sagittal and coronal slices. Despite the fact that the CNN was applied on 2D sagittal slices, almost no artifacts were visible on the synthetic CT and bone segmentation in the coronal plane (i.e. perpendicular to the direction the CNN was applied), demonstrating consistent performance across slices.

\begin{figure}
	\begin{center}
		\includegraphics[width=\textwidth]{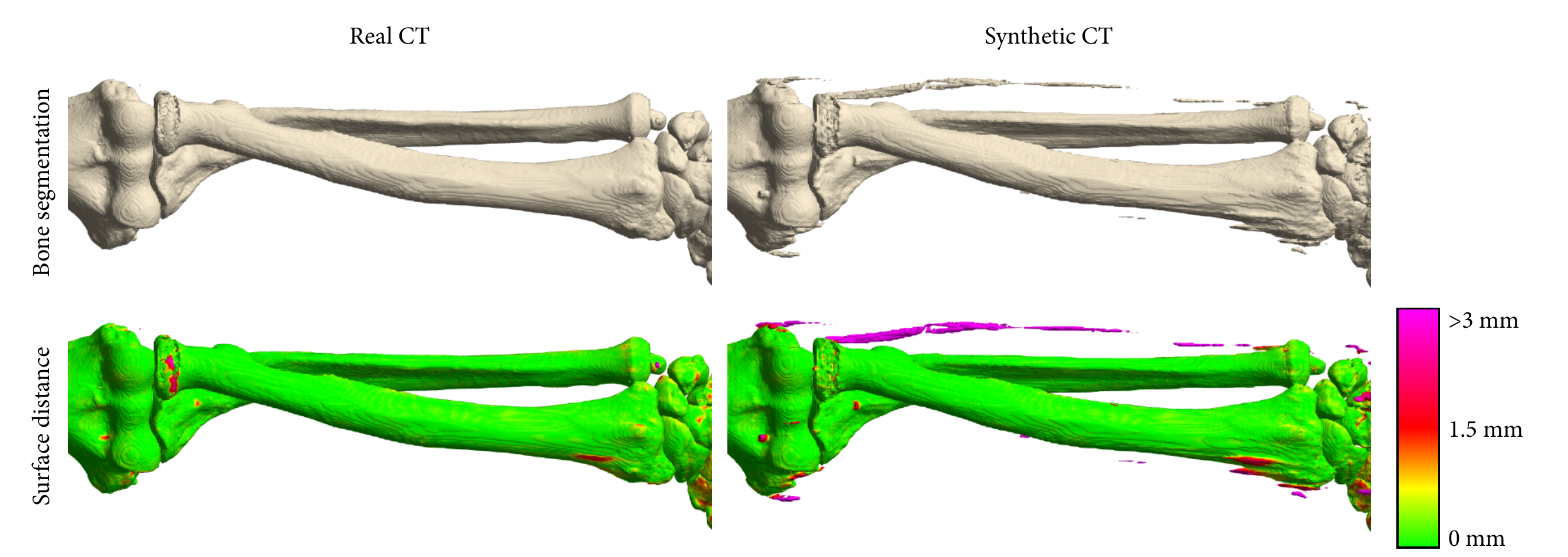}
	\end{center}
	
	\caption{Bone segmentations ($>200$ HU) and surface to surface distance maps for both real and synthetic CT for subject 3. The surface distance is the closest distance per point on the real CT surface to the synthetic CT surface (left) and vice versa (right). The distances were then projected as the colour on the bone segmentation.}
	\label{fig_render}
\end{figure}

Figure \ref{fig_render} shows bone segmentations and surface distance maps projected on both the real CT and the synthetic CT. In general the maps show that the shape of the bone on the synthetic CT closely matches that of the real CT. False positives sometimes occur on tendons, in particular when they are adjacent to bone. Since bone and tendons both show signal voids on MRI, distinguishing adjacent bone and tendon can be challenging. Other errors occur in the wrist, which is on the edge of the scan and may therefore not have optimal signal and could have been affected by magnetic field inhomogeneities. Furthermore, the field of view of the scans varied and did not always include the same amount of the wrist, which could have impeded training of the neural network on this area.

Table \ref{table} shows the quantitative metrics for all subjects. For all but subject 9 the metrics were in approximately the same range, which shows the results are consistent across the different folds in the cross validation and across subjects. The MRI scan for subject 9 showed a difference in contrast from the other 8 subjects, which could explain the poor performance. This difference may have been caused by the ex vivo tissues for this subject still being partially frozen.

\begin{table}
	\centering
	\vspace{5mm}
	\begin{tabular}{ r || c | c | c | c | c}
		\textbf{Subject} & \textbf{MAE overall (HU)} & \textbf{MAE bone (HU)} & \textbf{Dice bone} & \multicolumn{2}{l}{\textbf{Mean surface distance (mm)}} \\
		
		\rule{0pt}{1.1\normalbaselineskip} &&& & \textbf{sCT to CT} & \textbf{CT to sCT} \\
		\hline 
		\rule{0pt}{1.1\normalbaselineskip} 1 &	54.9 &	118.8 &	0.90 &	0.40 &	0.36 \\
		2 &	61.0 &	126.0 &	0.90 &	0.55 &	0.33 \\
		3 &	48.8 &	99.0 &	0.93 &	0.61 &	0.27 \\
		4 &	69.3 &	152.8 &	0.86 &	0.49 &	0.45 \\
		5 &	61.3 &	142.0 &	0.88 &	0.62 &	0.37 \\
		6 &	59.4 &	118.9 &	0.90 &	0.71 &	0.39 \\
		7 &	59.0 &	137.4 &	0.88 &	0.36 &	0.35 \\
		8 &	69.6 &	151.2 &	0.85 &	0.48 &	0.44 \\
		9 &	88.4 &	251.9 &	0.67 &	0.56 &	0.94 \\
		\hline
		\rule{0pt}{1.1\normalbaselineskip}  \textbf{Mean} &	63.5 &	144.2 &	0.86 &	0.53 &	0.43 \vspace{2mm} \\
	\end{tabular}
	
	\caption{Quantitative synthetic CT (sCT) evaluation metrics for all subjects. The Mean Absolute Error (MAE) is reported for both the overall tissue volume and for the bone. The Dice similarity and bidirectional mean surface distances are reported for the cortical bone segmentation ($>200$ HU).}
	\label{table}
\end{table}

The Dice similarity and mean surface distances indicate that general shape of the bone is well-represented in the synthetic CT. The false positives on tendons as shown in Figure 3 caused the synthetic CT to CT surface distance to be higher. The relatively high MAE values in the bone indicate that estimating exact Hounsfield units from MR images is challenging due to the lack of information on electron density in MRI.

\section{Discussion}
We have shown that high quality synthetic CT can be generated from multi-echo gradient echo MRI scans of the lower arm using a 2D conditional GAN neural network. Despite using very small training sets of only 4 subjects, the generated synthetic CT images showed good correspondence in bone segmentations and even reveal some of the structure in trabecular bone. Although tendons were sometimes confused for bone, the shape and structure of tendons could potentially be used to remove tendons from bone segmentations. Expanding the size of the training set may also help train the network to recognize these structures better.

The relatively high MAE values for the cortical bone indicate that the use of synthetic CT in applications that rely on quantitative interpretation of Hounsfield units may be challenging, for example in bone density measurements \cite{schreiber_use_2014}. However, the quality of the segmentation of the radius and ulna in particular shows that synthetic CT could be competitive with real CT in applications that depend on segmentations of these bones, such as osteotomy planning \cite{dobbe_computer-assisted_2011} and 3D printing \cite{auricchio_3d_2016}.

It is important to note that this study was limited to ex vivo scans in order to assess the quality of synthetic CT that can be expected to be generated from MR images. The lack of inter- and intra-scan motion in ex vivo scans simplified the CT to MR registration, which allowed the construction of a paired MR and CT dataset with negligible registration errors and corruption from motion artifacts. It is unlikely that a network trained on ex vivo data is directly applicable to in vivo scans, due to signal differences caused by physiology, such as blood flow, subject motion, and different tissue temperature. Creation of a paired in vivo dataset would involve a more difficult CT to MR registration problem because of potential rotation of the radius and ulna, which could negatively affect the synthetic CT quality. Recent developments with cycle-consistent GAN networks \cite{wolterink_deep_2017} could allow training using unregistered scans, but these networks will be harder to train stably and require more data than the cGAN approach presented in this study. The use of transfer learning to transfer networks trained on ex vivo scans to in vivo scans could prove helpful for adapting the methods to in vivo datasets.

In conclusion, this study shows that multi-echo gradient echo MR images contain sufficient information to allow accurate transformation to CT Hounsfield units for the purposes of bone segmentation. While these results on ex vivo scans are not likely to be directly applicable to in vivo scans, we do expect that these results can be replicated for in vivo scans, although the creation of a paired in vivo dataset will be more difficult. Under the condition that a high quality paired in vivo dataset can be created, MRI-based synthetic CT could be competitive with real CT for bone segmentation, and simultaneously provide additional information on soft tissues such as cartilage, muscles, tendons and ligaments.

\acknowledgments 

We gratefully acknowledge the support of NVIDIA Corporation with the donation of the Quadro P6000 GPU used for this research.

This work is part of the research programme Applied and Engineering Sciences (TTW) with project number 15479 which is (partly) financed by the Netherlands Organization for Scientific Research (NWO).

\bibliography{report} 
\bibliographystyle{spiebib} 

\end{document}